\pdfoutput=1

\documentclass[11pt]{article}

\usepackage[table,xcdraw]{xcolor}

\usepackage[final]{acl}

\usepackage{times}
\usepackage{latexsym}

\usepackage{amssymb}
\usepackage{multirow}
\usepackage{float}    
\usepackage{placeins} 
\usepackage{subcaption}  
\usepackage{wrapfig}
\usepackage{caption}  

\usepackage[T1]{fontenc}

\usepackage[utf8]{inputenc}

\usepackage{microtype}

\usepackage{inconsolata}

\usepackage{graphicx}

\usepackage{fancyhdr}
\usepackage{url}

\title{Jigsaw-Puzzles: From Seeing to Understanding to Reasoning in Vision-Language Models}

\author{
  \textbf{Zesen Lyu\textsuperscript{1,2}},
  \textbf{Dandan Zhang\textsuperscript{2}},
  \textbf{Wei Ye\textsuperscript{1}},
  \textbf{Fangdi Li\textsuperscript{2,3}},
  \textbf{Zhihang Jiang\textsuperscript{1,2}},
  \textbf{Yao Yang\textsuperscript{2}\thanks{Corresponding author.}}\\
  \textsuperscript{1}Hangzhou Institute for Advanced Study, UCAS \\
  \textsuperscript{2}Zhejiang Lab,
  \textsuperscript{3}Zhejiang University \\
  \{lvzesen23, yewei23, jiangzhihang23\}@mails.ucas.ac.cn, \\
  \{danae.zdd, yangyao\}@zhejianglab.com, FdbendyLi@zju.edu.cn
}

\begin{document}
\maketitle

\begin{abstract}
Spatial reasoning is a core component of human cognition, enabling individuals to perceive, comprehend, and interact with the physical world. It relies on a nuanced understanding of spatial structures and inter-object relationships, serving as the foundation for complex reasoning and decision-making. To investigate whether current vision-language models (VLMs) exhibit similar capability, we introduce Jigsaw-Puzzles, a novel benchmark consisting of 1,100 carefully curated real-world images with high spatial complexity. 
Based on this dataset, we design five tasks to rigorously evaluate VLMs’ spatial perception, structural understanding, and reasoning capabilities, while deliberately minimizing reliance on domain-specific knowledge to better isolate and assess the general spatial reasoning capability. 
We conduct a comprehensive evaluation across 24 state-of-the-art VLMs. The results show that even the strongest model, Gemini-2.5-Pro, achieves only 77.14\% overall accuracy and performs particularly poorly on the \textit{Order Generation} task, with only 30.00\% accuracy, far below the 90\%+ performance achieved by human participants. This persistent gap underscores the need for continued progress, positioning Jigsaw-Puzzles as a challenging and diagnostic benchmark for advancing spatial reasoning research in VLMs. Our project page is at \url{https://zesen01.github.io/jigsaw-puzzles}.
\end{abstract}

\section{Introduction}
\begin{figure}[!t]
 \centering
  \includegraphics[width=0.8\columnwidth]{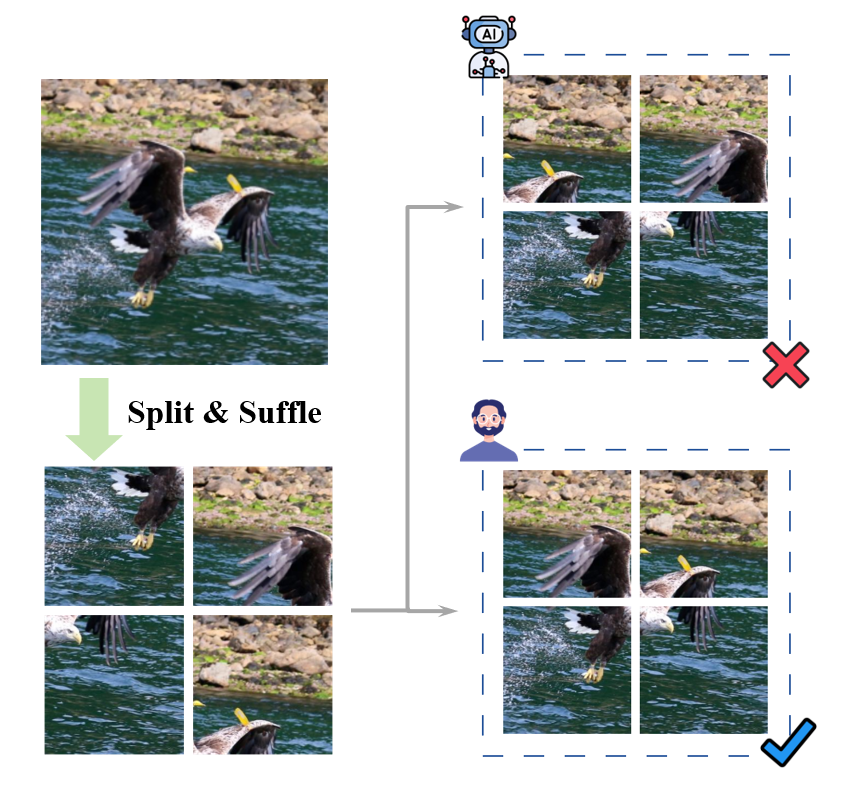}
  \caption{Jigsaw-Puzzles example. While human participants effortlessly reconstruct the original spatial layout, all tested VLMs fail to recover the correct order.}
  \label{fig:summary}
\end{figure}

\begin{figure}[!htbp]
\centering
  \includegraphics[width=1\columnwidth]{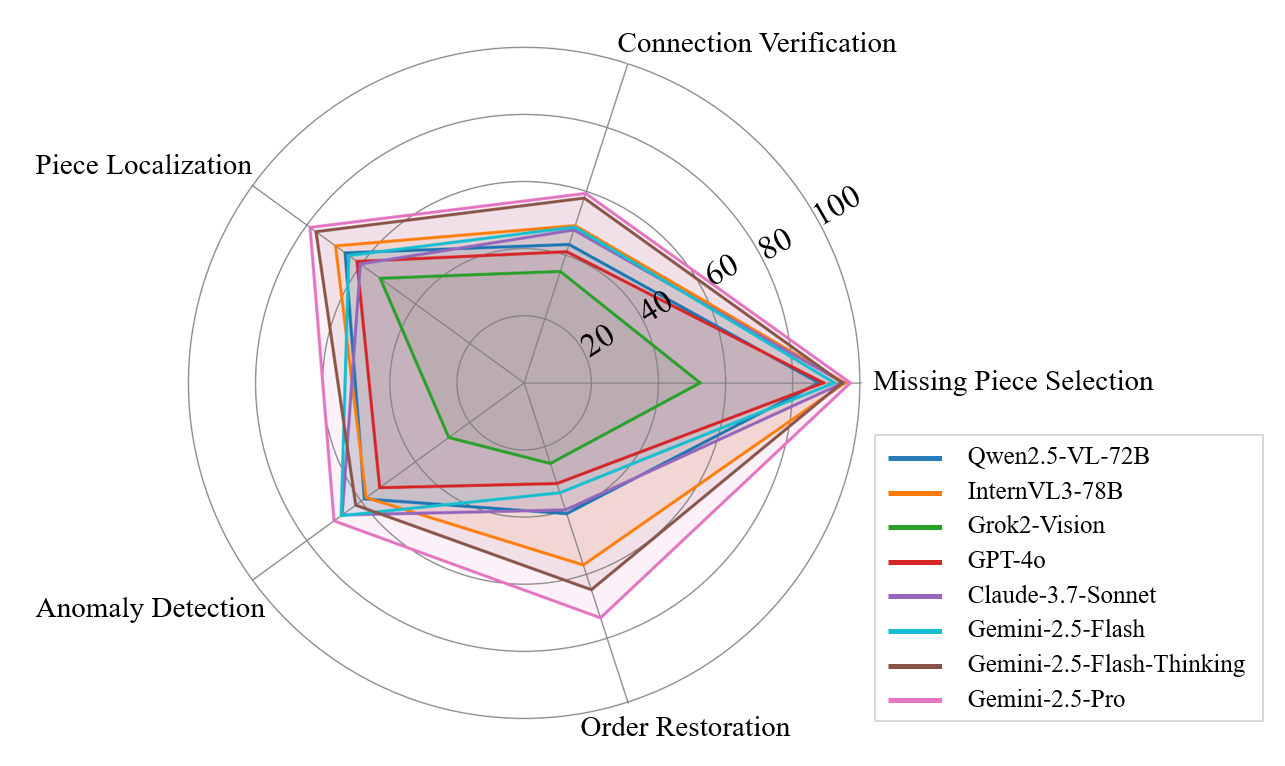}
  \caption{Evaluation of VLMs on Jigsaw-Puzzles. The plot reports the accuracy of 8 representative VLMs on 5 tasks.}
  \label{fig:radar}
\end{figure}

The road to artificial general intelligence (AGI) demands more than language or vision alone: it requires models to possess a human-like spatial reasoning capability by constructing structured representations of the physical world \cite{lake2017building}. Spatial reasoning refers not just to the perception of visual input, but to the capability to comprehend spatial arrangements, model structural relations, and infer the geometry and layout of a scene. These capabilities are fundamental to human cognition and develop naturally through everyday perception and interaction \cite{ishikawa2021spatial}. 
In contrast, current VLMs, while highly capable in tasks such as image captioning \cite{young2014image, lin2014microsoft, sharma2018conceptual}, visual question answering \cite{krishna2017visual, singh2019towards, marino2019ok}, and image-text retrieval \cite{schuhmann2021laion, thapliyal2022crossmodal, bitton2023breaking}, consistently struggle with tasks requiring spatial reasoning \cite{stogiannidis2025mind}. 
We show an example in Figure~\ref{fig:summary} and report the performance of some tested VLMs on Jigsaw-Puzzles in Figure~\ref{fig:radar}. 
This gap underscores a critical limitation: while current VLMs have made substantial progress in basic visual understanding, they continue to struggle with structured spatial reasoning, which is essential for grounded understanding in real-world scenarios. Bridging this gap is essential for progressing towards generalizable human-like spatial cognition and ultimately AGI.

However, existing benchmarks have yet to provide a comprehensive evaluation of spatial reasoning capability in VLMs under complex, real-world visual scenarios. 
Some works \cite{fu2024mmecomprehensiveevaluationbenchmark,li2024seed,liu2024mmbench,yue2024mmmu} focus primarily on foundational visual understanding by systematically evaluating perception, comprehension, and basic visual reasoning, revealing notable limitations in these areas. 
Although a few recent efforts \cite{pothiraj2025capture,stogiannidis2025mind,ren2025vgrp,tang2025lego} have attempted to evaluate the spatial reasoning capability of VLMs, they often rely on overly synthetic settings, task-specific constraints, or domain-dependent priors \cite{song2025visualpuzzles} (See Appendix~\ref{examples} for examples), limiting the capability to capture generalizable spatial reasoning under natural visual conditions. 
A truly effective evaluation of human-like spatial reasoning capability should model the task as a multi-stage cognitive process—beginning with perception, advancing through structural understanding, and culminating in high-level reasoning.
Such reasoning must be grounded in the visual richness and ambiguity of real images, requiring the integration of spatial structural modeling and goal-directed reasoning \cite{chen2024spatialvlm}. Yet, this critical dimension of spatial cognition remains largely overlooked in existing benchmarks, underscoring the need for new benchmarks that move beyond narrow task formulations and embrace the full complexity of spatial reasoning.

To overcome the limitations of existing benchmarks in evaluating spatial reasoning capability of VLMs under real-world conditions, we introduce Jigsaw-Puzzles, a novel benchmark inspired by the cognitive mechanisms underlying human puzzle-solving.  Puzzle-solving naturally reflects the multi-stage cognitive process \cite{fissler2018jigsaw}, making it a compelling testbed for spatial reasoning in VLMs. Unlike earlier studies that applied puzzle-solving primarily to visual representation learning \cite{noroozi2016unsupervised,wang2022video,shen2024satjip}, our benchmark employs it to evaluate the spatial reasoning ability of VLMs under realistic visual conditions.

In total, Jigsaw-Puzzles comprises 1,100 carefully curated real-world images and features five different tasks. 
First, we begin with the \textit{Missing Piece Selection} task to evaluate VLMs’ basic spatial understanding capability, which serves as the essential foundation for spatial reasoning.
Building on this foundation, we introduce four reasoning-centric tasks: \textit{Piece Localization}, \textit{Connection Verification}, \textit{Anomaly Detection}, and \textit{Order Restoration}. These tasks are designed to assess various facets of spatial reasoning, including adjacency modeling, local structural consistency, spatial localization, geometric transformation understanding, and multi-step spatial reasoning.

\begin{table}[]
\centering
\resizebox{\columnwidth}{!}{%
\begin{tabular}{lccccc}
\hline

 &  &  & {\color[HTML]{181A1C} } & {\color[HTML]{181A1C} } & {\color[HTML]{181A1C} } \\
\multirow{-2}{*}{\textbf{Benchmark}} & \multirow{-2}{*}{\textbf{Understanding}} & \multirow{-2}{*}{\textbf{Reasoning}} & \multirow{-2}{*}{{\color[HTML]{181A1C} \begin{tabular}[c]{@{}c@{}}\textbf{High Visual} \\ \textbf{Complexity}\end{tabular}}} & \multirow{-2}{*}{{\color[HTML]{181A1C} \begin{tabular}[c]{@{}c@{}}\textbf{Great} \\ \textbf{Scalability}\end{tabular}}} & \multirow{-2}{*}{{\color[HTML]{181A1C} \begin{tabular}[c]{@{}c@{}}\textbf{Automated}\\  \textbf{Construction}\end{tabular}}} \\ \hline

Capture \cite{pothiraj2025capture} & \checkmark & \checkmark & $\times$ & $\times$ & $\times$ \\
Mind the Gap \cite{stogiannidis2025mind} & \checkmark & \checkmark & $\times$ & $\times$ & $\times$ \\
VGRP \cite{ren2025vgrp} & \checkmark & \checkmark & $\times$ & $\times$ & $\times$ \\
LEGO-Puzzles \cite{tang2025lego} & \checkmark & \checkmark & $\times$ & \checkmark & \checkmark \\
\rowcolor[HTML]{FFFFC7} 
Jigsaw-Puzzles (Ours) & \textbf{\checkmark} & \textbf{\checkmark} & \textbf{\checkmark} & \textbf{\checkmark} & \textbf{\checkmark} \\ \hline
\end{tabular}
}
\caption{Comparison of spatial reasoning benchmarks.}
\label{tab:comparision}
\end{table}

Compared to existing benchmarks for evaluating spatial reasoning capability in VLMs, Jigsaw-Puzzles offers three key advantages, as summarized in Table~\ref{tab:comparision}: (1) Higher visual complexity. Jigsaw-Puzzles uses real-world images with diverse and rich visual elements, and significantly outperforms benchmarks based on synthetic images \cite{ren2025vgrp,stogiannidis2025mind,tang2025lego} and simple visual scenes \cite{pothiraj2025capture}. This enables more realistic and challenging spatial reasoning evaluation.
(2) Greater scalability. Any natural image that satisfies the construction rules can be directly used to generate puzzle tasks, without the need to manually synthesize target images.
(3) Fully automated construction pipeline. All Jigsaw-Puzzles tasks are generated automatically without manual annotation, with each question paired with a unique deterministic answer. This feature enables low-cost dataset construction and facilitates continuous expansion and refinement.

In summary, we introduce Jigsaw-Puzzles, a novel benchmark for systematically evaluating the human-like spatial reasoning capability of VLMs in realistic visual settings. Our main contributions are as follows:

\noindent \textbf{A new benchmark for spatial reasoning.} We introduce Jigsaw-Puzzles, a puzzle-inspired benchmark constructed through a fully automated pipeline that improves existing benchmarks in visual complexity and scalability, while enabling structured evaluation of spatial reasoning in VLMs.

\noindent \textbf{Comprehensive evaluation and analysis.} We evaluate 24 state-of-the-art VLMs on Jigsaw-Puzzles and conduct detailed analysis. Our findings expose consistent limitations in current VLMs and provide insights to guide future improvements in spatial reasoning capability.

\noindent \textbf{Open-sourced dataset and construction tools.} We release the full dataset along with the automated generation scripts to support the evaluation and continued advancement of spatial reasoning in VLMs under real-world visual scenarios, as well as to facilitate future benchmark expansion.

\section{Related Work}

\textbf{General VLMs Evaluation Benchmarks.} With the rapid progress of VLMs, systematically evaluating their diverse capabilities has become a key challenge. Although many benchmarks have been introduced, most focus primarily on visual understanding. MME \cite{fu2024mmecomprehensiveevaluationbenchmark} evaluates instruction following, perception, and basic reasoning across 14 subtasks, revealing persistent issues like object hallucination and limited spatial understanding. SEED-Bench \cite{li2024seed} includes 19,000 multiple-choice questions across 12 dimensions and shows continued struggles with text recognition and temporal reasoning. MMBench \cite{liu2024mmbench} offers fine-grained bilingual evaluations, enhancing the robustness of multilingual assessment. MMMU \cite{yue2024mmmu} provides 11,500 questions across 183 subfields and 30 image types to test expert-level reasoning, yet even advanced models like Gemini display notable knowledge gaps. While these benchmarks have advanced the evaluation of perceptual and semantic understanding, none systematically assess spatial reasoning—the core aspect of human cognition. This highlights the pressing need for more challenging and diagnostic benchmarks specifically targeting spatial reasoning capability in VLMs.

\noindent \textbf{Spatial Reasoning Evaluation Benchmarks in VLMs.} 
Several recent benchmarks have aimed to evaluate the spatial reasoning capability of VLMs. Capture \cite{pothiraj2025capture} assesses occluded object counting, revealing that VLMs struggle to form coherent spatial representations under occlusion. Mind the Gap \cite{stogiannidis2025mind} evaluates spatial relations, navigation, and mental rotation, showing that VLMs often perform near chance level, indicating limited spatial cognition. VGRP \cite{ren2025vgrp} introduces 20 visual grid puzzles across varying difficulty levels to assess visual perception, rule-following, and logical reasoning. LEGO-Puzzles \cite{tang2025lego} provides 1,100 visual QA pairs over 11 subtasks to measure basic and multi-step spatial reasoning. Results consistently show that current VLMs struggle with perceptual complexity, rotation reasoning, and sequential reasoning. Despite these efforts, most benchmarks rely on simplified scenarios, failing to reflect the complexity of real-world spatial environments, thereby limiting their generalizability. More diagnostic benchmarks grounded in natural visual settings are needed to advance human-level spatial reasoning in VLMs.

\section{Jigsaw-Puzzles}

In this section, we introduce Jigsaw-Puzzles, a scalable and comprehensive benchmark designed to evaluate the spatial reasoning capability of VLMs in realistic visual environments. Specifically, Section~\ref{sec:task_def} outlines the motivation and definition of each task, while Section~\ref{sec_data_cura} describes the dataset construction process, including image selection and the automated generation of question–answer pairs.

\subsection{Task Definition}
\label{sec:task_def}

\begin{figure*}[t]
  \centering
  \includegraphics[width=\textwidth]{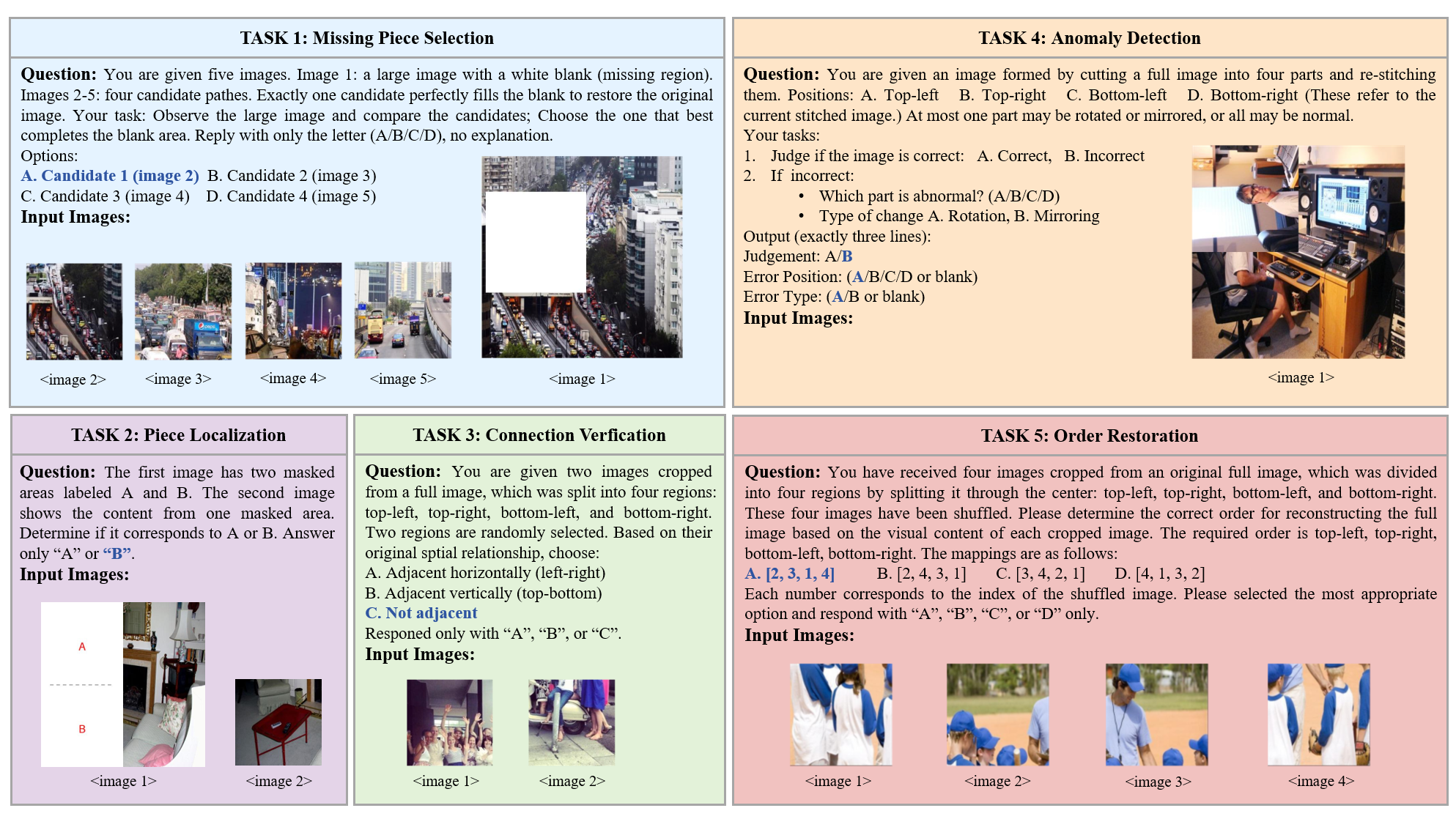}
  \caption{Task examples of Jigsaw-Puzzles. Note: the questions above are slightly simplified for clarity and brevity, and the \textcolor[HTML]{274b94}{\textbf{blue}} option indicates the correct answer.
}
  \label{fig:task_caption}
\end{figure*}

\begin{figure*}[!htbp]
  \centering
  \includegraphics[width=\textwidth]{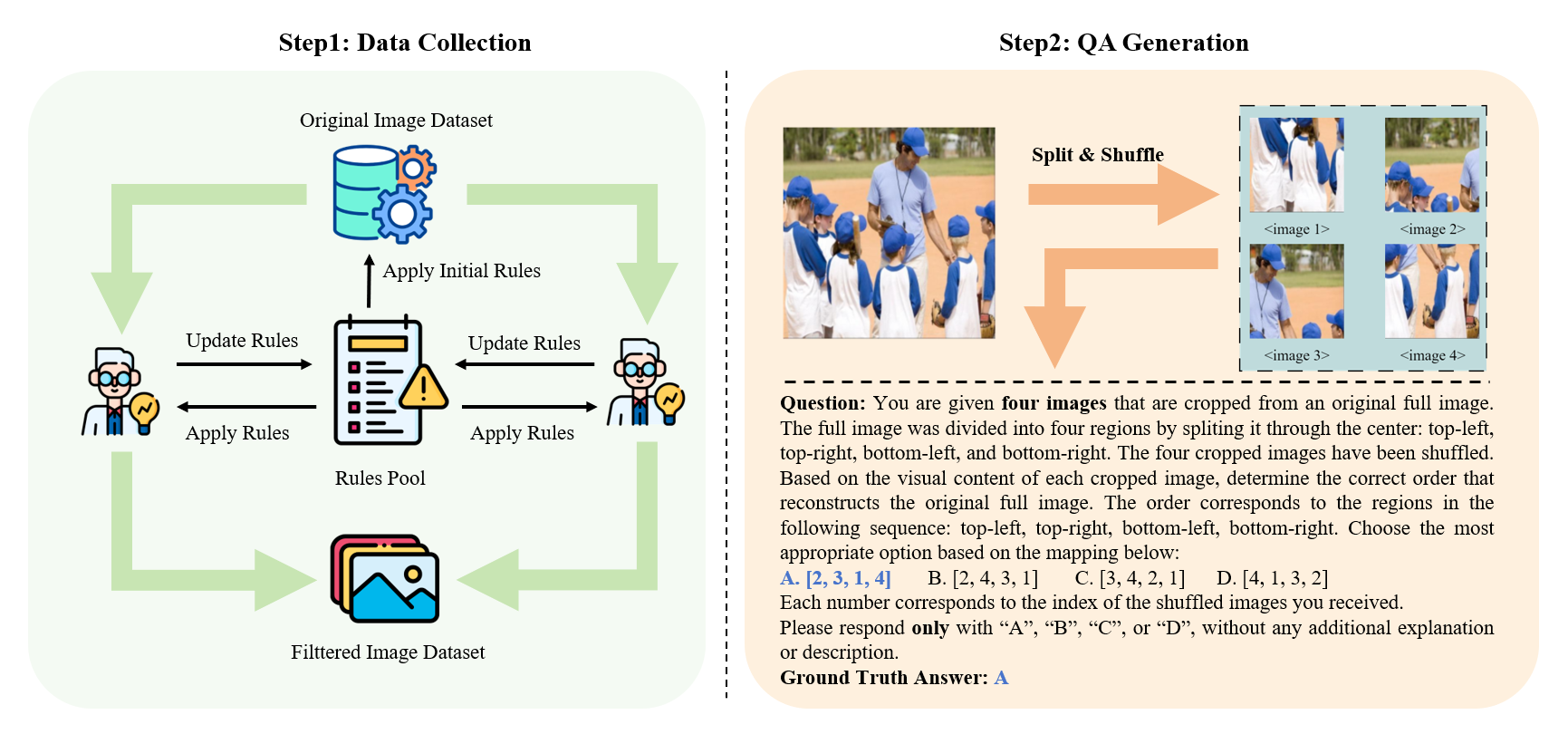}
  \caption{Dataset curation pipeline. Step 1 filters candidate images through expert-defined rules to build a spatial reasoning dataset. Step 2 uses automated templates to generate task-specific QA pairs from the curated images.}
  \label{fig:data_cura}
\end{figure*}

To systematically evaluate the spatial reasoning capability of VLMs, we design tasks around the core cognitive stages underlying human spatial reasoning—seeing, understanding, and reasoning. Inspired by the human process of solving jigsaw puzzles, our benchmark simulates how individuals integrate fragmented visual information into a coherent whole: beginning with the perception of local visual cues, followed by the comprehension of spatial relationships, and culminating in multi-step spatial reasoning to reconstruct the original scene. This sequence naturally reflects the progression from low-level perception to high-level spatial reasoning. Accordingly, the tasks span spatial understanding, single-step, and multi-step spatial reasoning, collectively providing a comprehensive evaluation across different levels of spatial reasoning. Figure~\ref{fig:task_caption} shows examples of each task in Jigsaw-Puzzles.

\textit{Task 1: Missing Piece Selection.} 
The task evaluates VLMs’ spatial understanding capability. Given an image with a missing region, VLMs need to select the correct patch from four candidates. We define two difficulty levels: Easy, where distractors are randomly chosen, and Hard, where distractors are selected using CLIP \cite{radford2021learning} similarity to closely resemble the ground-truth patch, increasing the task’s difficulty.
\textit{Task 2: Piece Localization}. This task evaluates spatial localization as a representative single-step spatial reasoning capability. Given a partially masked image and one masked patch, VLMs must identify the patch’s original position. Difficulty is controlled by grid size and number of masked regions: Easy (2×2 with two masks), Hard (3×3 with four masks), increasing spatial complexity.
\textit{Task 3: Connection Verification}. This task evaluates adjacency reasoning, which also falls under single-step spatial reasoning. The full image is divided into 2×2 grids, and two patches are randomly selected. VLMs are asked to determine their spatial relationship in the original image (e.g., above-below, left-right, or non-adjacent).
\textit{Task 4: Anomaly Detection}. This task targets local spatial transformation detection, a process that inherently involves single-step spatial reasoning. One region in 2×2 grids is randomly rotated, mirrored, or left unchanged. The model must detect the change, locate the region, and identify the transformation.
\textit{Task 5: Order Restoration}. This task integrates the capabilities assessed in the previous tasks and serves as a complex multi-step spatial reasoning challenge. A complete image is split into four shuffled patches. VLMs must identify the correct order to reconstruct the original spatial layout.

Overall, the five puzzle-inspired tasks in Jigsaw-Puzzles cover a broad spectrum of spatial reasoning challenges—from basic spatial understanding to single-step and multi-step spatial reasoning—enabling a comprehensive evaluation of spatial reasoning in VLMs.

\subsection{Dataset Curation}
\label{sec_data_cura}

As illustrated in Figure~\ref{fig:data_cura}, our dataset curation pipeline consists of two main stages: data collection and QA generation.

\noindent \textbf{Data Collection.} 
We integrate data collection and quality control into a unified process. Starting from the CC3M \cite{sharma2018conceptual} dataset, we apply task-specific filtering criteria—including minimum resolution and aspect ratio constraints—to construct an initial image pool of approximately 10,000 candidate images. Two human experts iteratively review the image pool while incrementally refining a shared set of filtering rules. Based on these evolving rules, they collaboratively filter the initial dataset to obtain the final set of high-quality, structurally diverse images. See Appendix~\ref{sec:dataset} for the rules pool.
To enhance generalizability, we emphasize semantic and structural diversity throughout the dataset. 

\noindent \textbf{QA Generation.}
To support scalable and consistent QA pairs generation, each task type is associated with a specific construction template. QA pairs are automatically generated using these templates. Figure~\ref{fig:task_caption} illustrates simplified examples of the templates, full versions are provided in Appendix~\ref{sec:dataset}.

\section{Evaluation on Jigsaw-Puzzles}

\subsection{Experimental Setting}

\textbf{Benchmark Models.} We evaluate 24 VLMs on Jigsaw-Puzzles, covering a diverse range of model scales and training paradigms. For open-source models, we evaluate Qwen2-VL-72B \cite{wang2024qwen2}, QvQ-72B-Preview \cite{qvq-72b-preview}, Qwen2.5-VL-[7B/32B/72B] \cite{bai2025qwen2}, InternVL3-[8B/14B/38B/78B] \cite{zhu2025internvl3}, Kimi-VL-A3B-[Instruct/Thinking] \cite{team2025kimi}, Phi-4-multimodal-instruct \cite{abouelenin2025phi}, Aya-Vision-[8B/32B] \cite{dash2025aya}, and Mistral-Small-3.1-24B-Instruct \cite{mistral}. For proprietary models, we evaluate Claude-[3.5/3.7]-Sonnet \cite{claude}, Gemini-[2.0/2.5]-Flash, Gemini-2.5-Flash-Thinking, Gemini-2.5-Pro \cite{team2023gemini}, GPT-4o, GPT-4o-mini \cite{achiam2023gpt}, and Grok-2-Vision \cite{grok}. Notably, QvQ-72B-Preview, Kimi-VL-A3B-Thinking, Gemini-2.5-Flash-Thinking, and Gemini-2.5-Pro are categorized as reasoning-enhanced models. All models, supporting multi-image input, are evaluated in a zero-shot setting with hardware scaled to their parameter size, see details in Appendix~\ref{sec:hardware}.

\begin{table*}[t]
\resizebox{\textwidth}{!}{%
\begin{tabular}{lcccccccr}
\hline
\multicolumn{1}{c}{} & \multicolumn{2}{c}{\cellcolor[HTML]{DAE9FD}\textbf{Missing Piece Selection}} & \multicolumn{2}{c}{\cellcolor[HTML]{FBE6CC}\textbf{Piece Localization}} & \cellcolor[HTML]{E1D6E7} & \cellcolor[HTML]{DCF4DC} & \cellcolor[HTML]{ECC4C2} & \multicolumn{1}{c}{} \\
\multicolumn{1}{c}{\multirow{-2}{*}{\textbf{Models}}} & \multicolumn{1}{>{\centering\arraybackslash}p{1.5cm}}{\textbf{Easy}} &
\multicolumn{1}{>{\centering\arraybackslash}p{1.5cm}}{\textbf{Hard}}  & \multicolumn{1}{>{\centering\arraybackslash}p{1.5cm}}{\textbf{Easy}} &
\multicolumn{1}{>{\centering\arraybackslash}p{1.5cm}}{\textbf{Hard}}  & \multirow{-2}{*}{\cellcolor[HTML]{E1D6E7}\textbf{\begin{tabular}[c]{@{}c@{}}Connection \\ Verification\end{tabular}}} & \multirow{-2}{*}{\cellcolor[HTML]{DCF4DC}\textbf{\begin{tabular}[c]{@{}c@{}}Anomaly\\ Detection\end{tabular}}} & \multirow{-2}{*}{\cellcolor[HTML]{ECC4C2}\textbf{\begin{tabular}[c]{@{}c@{}}Order\\ Restoration\end{tabular}}} & \multicolumn{1}{c}{\multirow{-2}{*}{\textbf{Overall}}} \\ \hline
\multicolumn{9}{l}{\cellcolor[HTML]{FFFFD1}Baseline} \\
Random Guessing & 25.00 & 25.00 & 50.00 & 25.00 & 33.33 & 28.13 & \multicolumn{1}{c|}{25.00} & 30.21 \\
↑ Random (p < 0.05) & 27.30 & 27.30 & 52.50 & 27.30 & 35.70 & 30.50 & \multicolumn{1}{c|}{27.30} & 32.56 \\ \hline
\multicolumn{9}{l}{\cellcolor[HTML]{FFFFD1}Proprietary   Models} \\
Grok2-Vision & 64.55 & 52.45 & 53.00 & 41.00 & 34.91 & 27.73 & \multicolumn{1}{c|}{25.27} & 42.70 \\
GPT-4o-mini & 96.45 & 83.64 & 59.45 & 37.82 & 44.36 & 57.91 & \multicolumn{1}{c|}{33.18} & 58.97 \\
GPT-4o & 95.00 & 89.18 & 61.55 & 53.45 & 41.09 & 53.18 & \multicolumn{1}{c|}{31.55} & 60.71 \\
Claude-3.5-Sonnet & \cellcolor[HTML]{CBF7CA}99.73 & 94.55 & 62.45 & 41.09 & 45.64 & 67.45 & \multicolumn{1}{c|}{35.00} & 63.70 \\
Claude-3.7-Sonnet & 99.55 & \cellcolor[HTML]{CBF7CA}95.09 & 60.27 & 44.55 & 47.91 & 67.00 & \multicolumn{1}{c|}{39.82} & 64.88 \\
Gemini-2.0-Flash & 92.09 & 85.64 & 63.55 & 54.00 & 44.91 & \cellcolor[HTML]{CBF7CA}68.73 & \multicolumn{1}{c|}{34.27} & 63.31 \\
Gemini-2.5-Flash & 98.82 & 92.45 & 64.55 & \cellcolor[HTML]{CBF7CA}54.55 & 48.82 & 67.36 & \multicolumn{1}{c|}{34.45} & \cellcolor[HTML]{B7FFFB}65.86 \\
\textbf{Gemini-2.5-Flash-Thinking} & \textbf{99.55} & \textbf{94.73} & \cellcolor[HTML]{CBF7CA}\textbf{76.64} & \textbf{51.27} & \cellcolor[HTML]{CBF7CA}\textbf{57.91} & \textbf{62.00} & \multicolumn{1}{c|}{\cellcolor[HTML]{CBF7CA}\textbf{64.82}} & \cellcolor[HTML]{36FFF4}\textbf{72.42} \\
\textbf{Gemini-2.5-Pro} & \cellcolor[HTML]{34FF34}\textbf{99.91} & \cellcolor[HTML]{34FF34}\textbf{97.18} & \cellcolor[HTML]{34FF34}\textbf{78.82} & \cellcolor[HTML]{34FF34}\textbf{61.09} & \cellcolor[HTML]{34FF34}\textbf{59.36} & \cellcolor[HTML]{34FF34}\textbf{70.00} & \multicolumn{1}{c|}{\cellcolor[HTML]{34FF34}\textbf{73.64}} & \cellcolor[HTML]{36D1C8}\textbf{77.14} \\ \hline
\multicolumn{9}{l}{\cellcolor[HTML]{FFFFD1}Open-source   Models} \\
Kimi-VL-A3B-Instruct & 67.91 & 52.55 & 51.45 & 29.82 & 37.91 & 21.82 & \multicolumn{1}{c|}{32.82} & 42.04 \\
\textbf{Kimi-VL-A3B-Thinking} & \textbf{84.64} & \textbf{58.09} & \textbf{56.36} & \textbf{32.64} & \textbf{28.00} & \textbf{30.91} & \multicolumn{1}{c|}{\textbf{20.36}} & \textbf{44.43} \\
Phi-4-multimodal-instruct & 63.45 & 51.64 & 60.91 & 37.45 & 36.64 & 43.36 & \multicolumn{1}{c|}{27.64} & 45.87 \\
Qwen2.5-VL-7B & 87.18 & 63.27 & 54.27 & 36.18 & 38.55 & 45.00 & \multicolumn{1}{c|}{28.09} & 50.36 \\
Aya-Vision-8B & 26.82 & 27.27 & 49.64 & 26.55 & 35.00 & 12.91 & \multicolumn{1}{c|}{24.73} & 28.99 \\
InternVL3-8B & 98.09 & 85.45 & 53.91 & 35.45 & 44.91 & 46.82 & \multicolumn{1}{c|}{34.36} & 57.00 \\
InternVL3-14B & \cellcolor[HTML]{34FF34}99.73 & 88.73 & 59.64 & 40.18 & \cellcolor[HTML]{CBF7CA}51.73 & 49.09 & \multicolumn{1}{c|}{40.91} & 61.43 \\
Mistral-Small-3.1-24B-Instruct & 26.91 & 28.55 & 54.27 & 31.27 & 38.27 & 52.36 & \multicolumn{1}{c|}{26.45} & 36.87 \\
Qwen2.5-VL-32B & 97.27 & 76.82 & 62.09 & 40.36 & 50.09 & \cellcolor[HTML]{34FF34}61.55 & \multicolumn{1}{c|}{33.64} & 60.26 \\
Aya-Vision-32B & 24.27 & 25.27 & 51.45 & 28.36 & 37.18 & 41.45 & \multicolumn{1}{c|}{25.00} & 33.28 \\
InternVL3-38B & 99.00 & \cellcolor[HTML]{CBF7CA}91.36 & 63.45 & 42.64 & \cellcolor[HTML]{34FF34}56.73 & 30.73 & \multicolumn{1}{c|}{\cellcolor[HTML]{CBF7CA}54.64} & \cellcolor[HTML]{B7FFFB}62.65 \\
Qwen2-VL-72B & 95.55 & 75.36 & 55.27 & 40.64 & 40.55 & 42.36 & \multicolumn{1}{c|}{33.55} & 54.75 \\
\textbf{QVQ-72B-Preview} & \textbf{77.82} & \textbf{52.18} & \textbf{53.09} & \textbf{36.73} & \textbf{41.82} & \textbf{47.73} & \multicolumn{1}{c|}{\textbf{34.64}} & \textbf{49.14} \\
Qwen2.5-VL-72B & \cellcolor[HTML]{CBF7CA}99.36 & 87.82 & \cellcolor[HTML]{CBF7CA}65.91 & \cellcolor[HTML]{CBF7CA}45.27 & 43.36 & \cellcolor[HTML]{CBF7CA}58.82 & \multicolumn{1}{c|}{41.00} & \cellcolor[HTML]{36FFF4}63.08 \\
InternVL3-78B & \cellcolor[HTML]{34FF34}99.73 & \cellcolor[HTML]{34FF34}95.55 & \cellcolor[HTML]{34FF34}69.45 & \cellcolor[HTML]{34FF34}52.27 & 49.27 & 58.18 & \multicolumn{1}{c|}{\cellcolor[HTML]{34FF34}57.09} & \cellcolor[HTML]{36D1C8}68.79 \\ \hline
\end{tabular}%
}
\caption{Full Evaluation Results of 24 VLMs on Jigsaw-Puzzles. VLMs are grouped into proprietary and open-source categories. 
\colorbox[HTML]{34FF34}{Dark Green} and \colorbox[HTML]{CBF7CA}{Light Green} indicate the top-1 and top-2 performance within each group, respectively. Results of reasoning-enhanced are marked in \textbf{bold}. We also highlight the top three models based on their overall performance, using \colorbox[HTML]{36D1C8}{Dark Blue}, \colorbox[HTML]{36FFF4}{Medium Blue}, and \colorbox[HTML]{B7FFFB}{Light Blue}, respectively.
}
\label{tab:all_results}
\end{table*}

\begin{table*}[!htbp]
\resizebox{\textwidth}{!}{%
\begin{tabular}{lcccccccr}
\hline
\multicolumn{1}{c}{} & \multicolumn{2}{c}{\cellcolor[HTML]{DAE9FD}\textbf{Missing Piece Selection}} & \multicolumn{2}{c}{\cellcolor[HTML]{FBE6CC}\textbf{Piece Localization}} & \cellcolor[HTML]{E1D6E7} & \cellcolor[HTML]{CADDCA} & \cellcolor[HTML]{ECC4C2} & \multicolumn{1}{c}{} \\
\multicolumn{1}{c}{\multirow{-2}{*}{\textbf{Models}}} & \multicolumn{1}{>{\centering\arraybackslash}p{1.5cm}}{\textbf{Easy}} &
\multicolumn{1}{>{\centering\arraybackslash}p{1.5cm}}{\textbf{Hard}}  & \multicolumn{1}{>{\centering\arraybackslash}p{1.5cm}}{\textbf{Easy}} &
\multicolumn{1}{>{\centering\arraybackslash}p{1.5cm}}{\textbf{Hard}} & \multirow{-2}{*}{\cellcolor[HTML]{E1D6E7}\textbf{\begin{tabular}[c]{@{}c@{}}Connection \\ Verification\end{tabular}}} & \multirow{-2}{*}{\cellcolor[HTML]{CADDCA}\textbf{\begin{tabular}[c]{@{}c@{}}Anomaly\\ Detection\end{tabular}}} & \multirow{-2}{*}{\cellcolor[HTML]{ECC4C2}\textbf{\begin{tabular}[c]{@{}c@{}}Order \\ Restoration\end{tabular}}} & \multicolumn{1}{c}{\multirow{-2}{*}{\textbf{Overall}}} \\ \hline
Human   Performance & \cellcolor[HTML]{34FF34}99.55 & \cellcolor[HTML]{34FF34}100.00 & \cellcolor[HTML]{34FF34}95.45 & \cellcolor[HTML]{34FF34}91.36 & \cellcolor[HTML]{34FF34}93.18 & \cellcolor[HTML]{34FF34}97.27 & \multicolumn{1}{c|}{\cellcolor[HTML]{34FF34}97.73} & \cellcolor[HTML]{36D1C8}96.36 \\ \hline
\multicolumn{9}{l}{\cellcolor[HTML]{FFFFD1}\textbf{Proprietary Models}} \\
Claude-3.7-Sonnet & 100.00 & 95.45 & 55.45 & 47.27 & 42.73 & 68.18 & \multicolumn{1}{c|}{38.64} & 63.96 \\
Gemini-2.5-Flash & 98.18 & 93.18 & 58.18 & 55.45 & 42.27 & 66.82 & \multicolumn{1}{c|}{32.27} & 63.76 \\
\textbf{Gemini-2.5-Flash-Thinking} & \textbf{99.55} & \textbf{95.91} & \textbf{71.82} & \textbf{51.82} & \textbf{55.00 }& \textbf{60.91} & \multicolumn{1}{c|}{\textbf{57.27}} & \textbf{\cellcolor[HTML]{B7FFFB}70.33} \\
\textbf{Gemini-2.5-Pro} & \textbf{100.00 }& \textbf{96.36} & \textbf{77.73} & \textbf{56.82} & \textbf{57.27} & \textbf{71.36} & \multicolumn{1}{c|}{\textbf{70.91}} & \textbf{\cellcolor[HTML]{36FFF4}75.78} \\ \hline
\multicolumn{9}{l}{\cellcolor[HTML]{FFFFD1}\textbf{Open-source Models}} \\
Qwen2.5-VL-72B & 99.09 & 86.82 & 67.73 & 42.27 & 40.00 & 57.73 & \multicolumn{1}{c|}{33.64} & 61.04 \\
InternVL3-78B & 99.55 & 95.45 & 70.45 & 53.64 & 44.55 & 61.36 & \multicolumn{1}{c|}{56.82} & 68.83 \\ \hline
\end{tabular}%
}
\caption{Comparing Top-Performing VLMs with Human Performance on Jigsaw-Puzzles-Lite. The human performance is highlighted in \colorbox[HTML]{34FF34}{Dark Green}. Results of reasoning-enhanced are marked in \textbf{bold}. 
The top three overall performance are highlighted in \colorbox[HTML]{36D1C8}{Dark Blue}, \colorbox[HTML]{36FFF4}{Medium Blue}, and \colorbox[HTML]{B7FFFB}{Light Blue}, respectively.
}
\label{tab:human}
\end{table*}

\noindent \textbf{Evaluation Metric.} Since each QA pair in Jigsaw-Puzzles has a single correct answer, we use exact match accuracy (\%) as the primary metric to evaluate VLMs’ performance on each task.

\noindent \textbf{Baselines.} We provide two baselines for comparison: (1) Random, which assumes equal probability across all options and calculates expected accuracy accordingly. (2) p-value-based critical value, which reports the minimum accuracy required to outperform random guessing at a significance level of p=0.05.

\noindent \textbf{Human Performance.} To evaluate human performance, we construct a subset called Jigsaw-Puzzles-Lite by sampling 220 images from the full dataset. 
Three human participants complete all tasks on this subset under the same conditions as VLMs—without access to any external tools or the internet. Their performance serves as an empirical upper bound for spatial reasoning capability.

\subsection{Main Results}

Table~\ref{tab:all_results}, \ref{tab:human} report the performance of 24 VLMs on Jigsaw-Puzzles. Building on these results, we conduct a comprehensive and systematic analysis. We summarize several key findings as below.

\noindent \textbf{Spatial Reasoning Remains a Challenge for VLMs.} As shown in Table~\ref{tab:human}, human participants consistently outperform VLMs, achieving an overall accuracy of 96.36\%. By comparison, current VLMs perform considerably worse, with even the strongest models—Gemini-2.5-Pro—lagging more than 20 percentage points behind human accuracy across all tasks.
The persistent gap between humans and VLMs highlights the demanding nature of Jigsaw-Puzzles and affirms its utility as a robust benchmark for spatial reasoning evaluation.

\noindent \textbf{Significant Gap Between Open-Source and Proprietary VLMs.} 
As shown in Tables~\ref{tab:all_results}, proprietary VLMs consistently outperform open-source VLMs on Jigsaw-Puzzles. 
Among them, non-reasoning-enhanced proprietary VLMs typically exceed 60\% overall accuracy, whereas most open-source VLMs fall short—only InternVL3-[14B/38B/78B] and Qwen2.5-VL-72B surpass this threshold. Reasoning-enhanced proprietary models, such as Gemini-2.5-Flash-Thinking (72.42\%) and Gemini-2.5-Pro (77.14\%), further widen this gap. 
These results reveal a persistent disparity in spatial reasoning performance, suggesting that proprietary VLMs benefit from advantages in model architecture, training strategy, and access to large-scale data. Meanwhile, this finding highlights substantial room for improvement in open-source VLMs toward achieving more robust and generalizable spatial reasoning.

\noindent \textbf{Model Performance in Different Tasks.} 
In the \textit{Missing Piece Selection} task, which primarily targets spatial understanding, most proprietary VLMs perform well under both Easy and Hard settings, demonstrating strong perceptual capability. Although open source models generally perform poorly by comparison, certain models, such as the InternVL3 series and Qwen2.5-VL-72B, achieve perceptual understanding on par with proprietary VLMs. Notably, both the Aya-Vision series and the Mistral-Small-3.1-24B-Instruct models perform poorly across all settings, even at the 32B scale, accuracy remains near random, revealing severe deficits in spatial understanding and instruction following.
In single-step spatial reasoning tasks—\textit{Piece Localization}, \textit{Connection Verification},  and \textit{Anomaly Detection}—most VLMs surpass the p-value-based critical value, indicating emerging competence in basic spatial reasoning. However, strong performance remains concentrated in only a few models, particularly reasoning-enhanced proprietary models and the latest open-source InternVL3 series. This disparity becomes even more evident in the multi-step spatial reasoning task—\textit{Order Restoration}, indicating that most VLMs struggle with complex spatial reasoning.

\begin{figure}[t]
  \includegraphics[width=0.9\columnwidth]{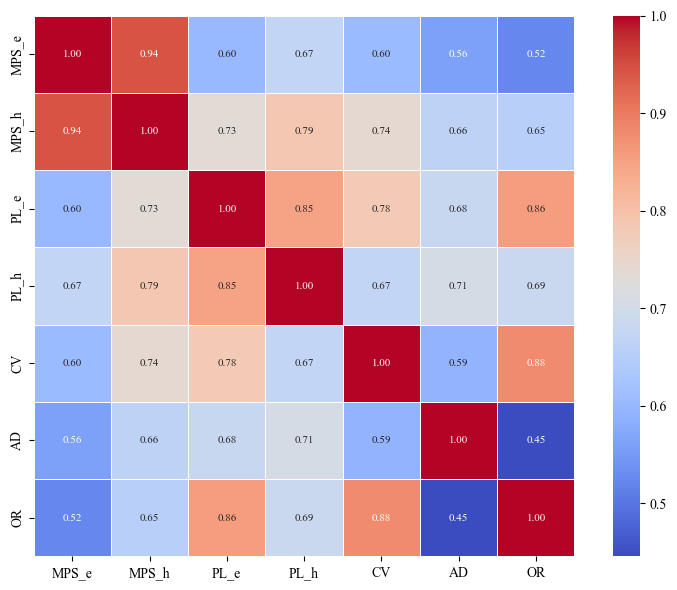}
  \centering
  \caption{Task Similarity Heatmap. The heatmap illustrates the pairwise correlation between tasks in our benchmark, measured using Pearson correlation coefficients. Task names are abbreviated using the initials of each word (e.g., \textit{Missing Piece Selection} → MPS). The suffixes \_e and \_h indicate the Easy and Hard settings, respectively.}
  \label{fig:similar}
\end{figure}

\begin{figure}[!htbp]
  \includegraphics[width=\columnwidth]{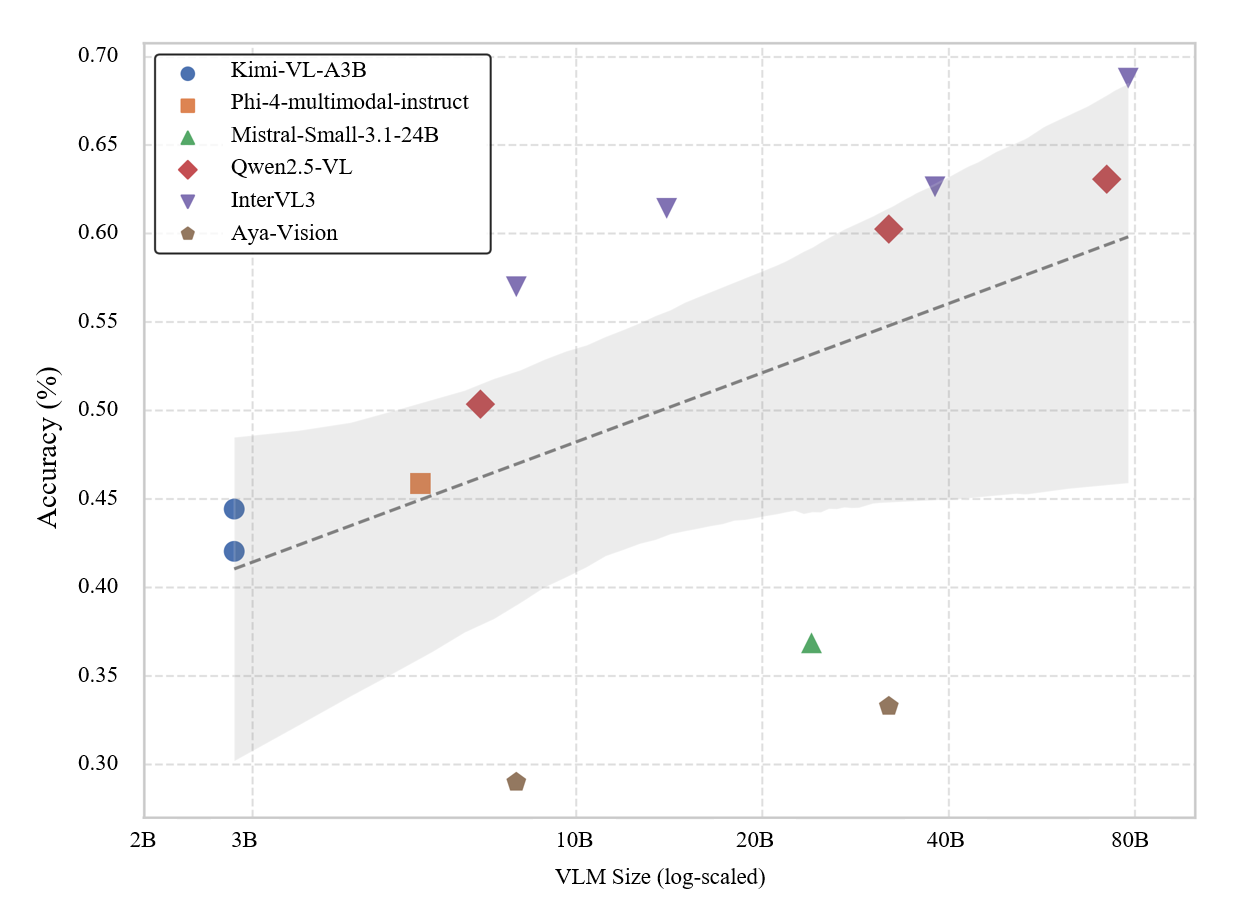}
  \caption{Relationship between VLM size and performance on Jigsaw-Puzzles. Each point represents a VLM, with its accuracy plotted against log-scaled parameter size. A clear positive correlation is observed both across and within model families, indicating that larger models tend to exhibit stronger performance.}
  \label{fig:scaling}
\end{figure}

In conclusion, Jigsaw-Puzzles effectively distinguishes VLMs across a spectrum of spatial reasoning capability—from basic understanding to complex multi-step reasoning. As shown by the results in Table~\ref{tab:all_results}, substantial room for improvement remains, particularly in multi-step spatial reasoning. 

\noindent \textbf{Foundational Spatial Understanding Shapes Reasoning Performance.}
We analyze task similarity on Jigsaw-Puzzles by computing Pearson correlation coefficients between each task and all others, as proposed by \citet{zhang2025redundancy}, as shown in Figure~\ref{fig:similar}. The results show that performance on the \textit{Missing Piece Selection} task—a proxy for spatial understanding, is strongly correlated with performance on spatial reasoning tasks. In contrast, VLMs with weaker spatial understanding often struggle with reasoning tasks, with some performing worse than random on reasoning-intensive tasks. This pattern reflects the human cognitive progression from perception to understanding to reasoning, underscoring the foundational role of spatial understanding in enabling higher-level spatial reasoning in VLMs.

\noindent \textbf{Spatial Reasoning Scales with VLM size.}
We analyze the relationship between VLM size and overall performance on Jigsaw-Puzzles. As shown in Figure~\ref{fig:scaling}, our results reveal that VLM accuracy consistently increases with model size, both across all models and within specific families (e.g., InternVL3, Qwen2.5-VL). This positive correlation suggests that spatial reasoning capability—like other cognitive competencies \cite{wang2024coglm}—benefits from larger model capacity, which scales with parameter count.

\begin{figure}[t]
\centering
  \includegraphics[width=0.9\columnwidth]{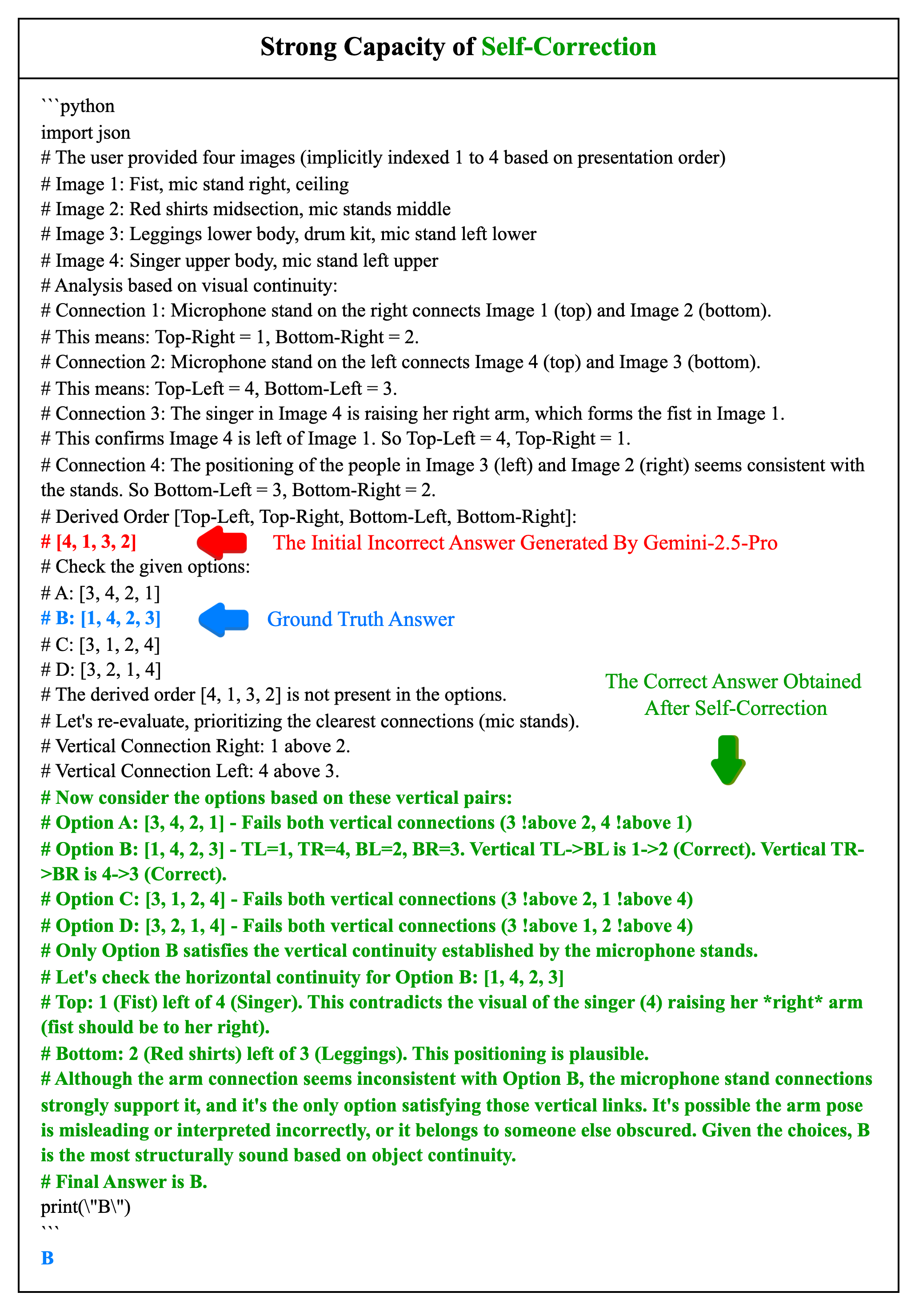}
  \caption{An example of self-correction. \textcolor[HTML]{ff0000}{Red} shows the initial incorrect answer generated by Gemini-2.5-Pro; \textcolor[HTML]{0074ff}{Blue} indicates the ground-truth answer; \textcolor[HTML]{008e00}{Green} illustrates the model’s self-correction process.}
  \label{fig:self_correct}
\end{figure}

\noindent \textbf{Reasoning-Enhanced Models Show Superior Spatial Reasoning Performance.} To assess the spatial reasoning capability of reasoning-enhanced VLMs, we evaluate Gemini-2.5-Flash-Thinking, Gemini-2.5-Pro, Kimi-VL-A3B-Thinking and QvQ-72B-Preview. Except for Gemini-2.5-Pro, each model has a corresponding base version for comparison. As shown in Table~\ref{tab:all_results}, these enhanced VLMs consistently achieve higher overall accuracy. For example, Kimi-VL-A3B-Thinking improves from 42.04\% to 44.43\%, and Gemini-2.5-Flash-Thinking rises from 65.86\% to 72.42\%. Although QvQ-72B-Preview overall underperforms Qwen2-VL-72B, it achieves better results on spatial reasoning tasks. Notably, Gemini-2.5-Pro achieves the highest overall accuracy (77.14\%) among all VLMs tested.
Furthermore, the largest improvements occur in the multi-step spatial reasoning task, \textit{Order Restoration}, where reasoning-enhanced VLMs outperform their base counterparts more substantially than in single-step tasks. To explain this, we analyze cases where only Gemini-2.5-Pro answers correctly, with Figure~\ref{fig:self_correct} presenting one such example. Gemini-2.5-Pro demonstrates a form of self-correction: when the model’s initial prediction is not among the provided options, it will re-evaluate the visual input and revise its judgment. This behavior, facilitated by the reduced answer space under choice constraints, may contribute to the superior performance of reasoning-enhanced models in the \textit{Order Restoration} task.

\begin{figure}[t]
\centering
  \includegraphics[width=0.9\columnwidth]{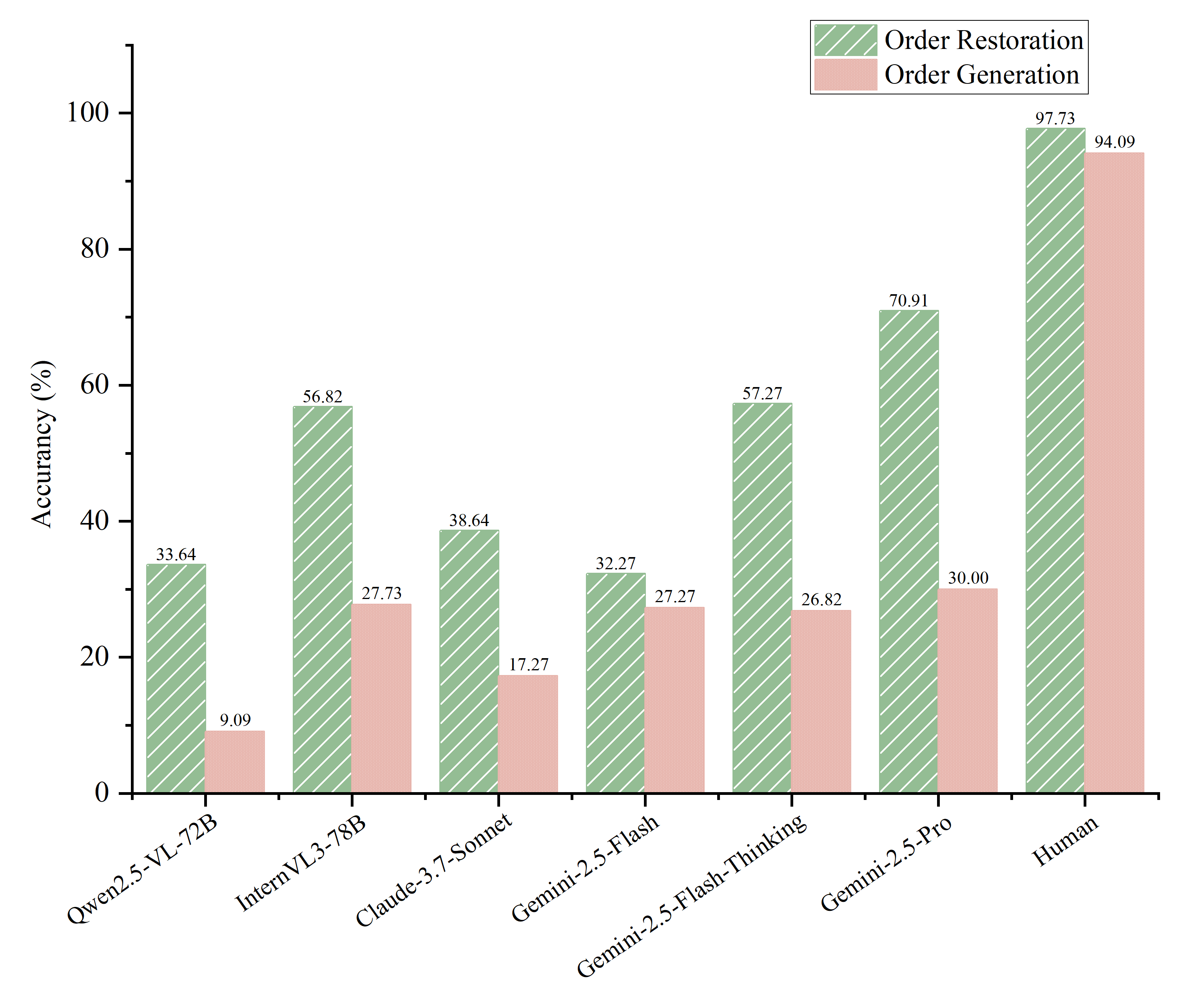}
  \caption{Evaluation of \textit{Order Restoration} and \textit{Order Generation} tasks on Jigsaw-Puzzles-Lite. Without option constraints, VLM accuracy drops significantly—peaking at just 30.00\% and falling far short of human performance.}
  \label{fig:genrate}
\end{figure}

\noindent \textbf{Further Exploring Multi-Step Spatial Reasoning in VLMs.} 
To further evaluate VLMs’ multi-step spatial reasoning beyond the constraints of predefined choices, we introduce the \textit{Order Generation} task based on Jigsaw-Puzzles-Lite. In this setting, VLMs must directly generate the correct sequence of puzzle pieces without relying on answer options, thereby more authentically simulating open-ended spatial reasoning. As shown in Figure~\ref{fig:genrate}, current VLMs consistently struggle with this task—Gemini-2.5-Pro, the best-performing model, achieves only 30.00\% accuracy, in stark contrast to 94.09\% by human participants. This finding reveals that, despite exhibiting strong self-correction behavior under option constraints, existing VLMs face considerable challenges in autonomously constructing coherent spatial reasoning chains. This highlights a significant gap between current VLMs and human-level spatial reasoning in open-ended scenarios.

\section{Conclusion}

We propose Jigsaw-Puzzles, a novel benchmark for systematically evaluating the spatial reasoning capability of VLMs in real-world visual scenes. Through extensive experiments on 24 representative VLMs, we identify persistent gaps between current VLMs and human-level spatial reasoning—especially in multi-step spatial reasoning tasks. Jigsaw-Puzzles provides a scalable and cognitively grounded benchmark to advance future research on spatial reasoning in VLMs.

\section*{Limitations}

While Jigsaw-Puzzles provides a structured benchmark tailored for 2D spatial reasoning in static images, it currently does not address 3D perception, temporal sequences, or embodied contexts—each of which represents an important and orthogonal axis of spatial cognition. We view this as a natural next step and encourage future work to extend the benchmark in these directions.

\section*{Acknowledgments}

The authors wish to thank the anonymous reviewers for their helpful comments.
This work was supported by the Key R\&D Program of Zhejiang (2024C01036).

\bibliography{custom}

\appendix

\section{Examples of Other Spatial Reasoning Benchmarks}
\label{examples}

\begin{figure}[H]
\centering
  \includegraphics[width=\columnwidth]{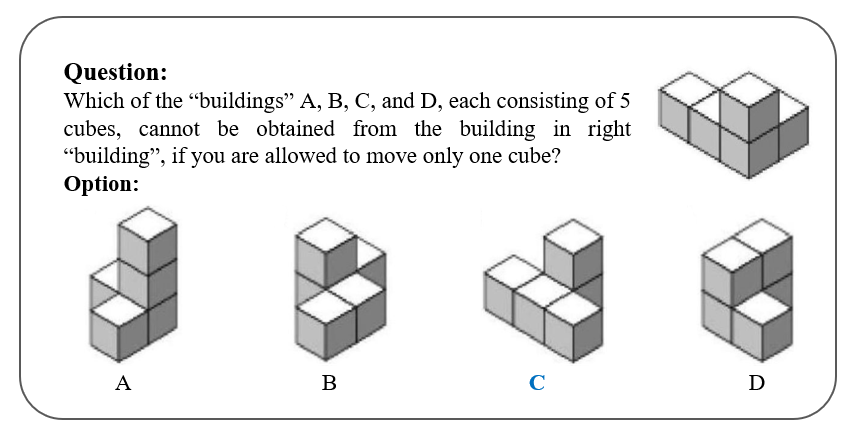}
  \caption{An example of Mind the Gap \cite{stogiannidis2025mind}.}
  \label{fig:cube}
\end{figure}

\begin{figure}[H]
\centering
  \includegraphics[width=\columnwidth]{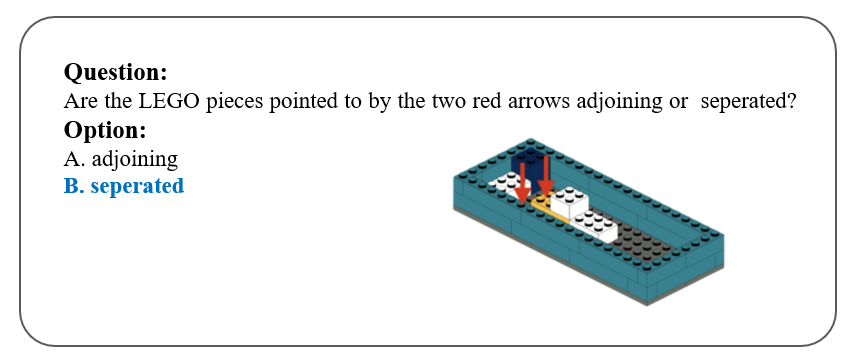}
  \caption{An example of LEGO-Puzzles \cite{tang2025lego}.}
  \label{fig:lego}
\end{figure}

Figure~\ref{fig:cube} and Figure~\ref{fig:lego} illustrate two representative question types commonly used to evaluate spatial reasoning capability of VLMs. The tested images are not based on real-world scenes, which limits the capability to evaluate spatial reasoning in VLMs under realistic conditions.

\FloatBarrier

\section{Dataset Curation}
\label{sec:dataset}

\begin{figure}[!htbp]
\centering
  \includegraphics[width=0.9\columnwidth]{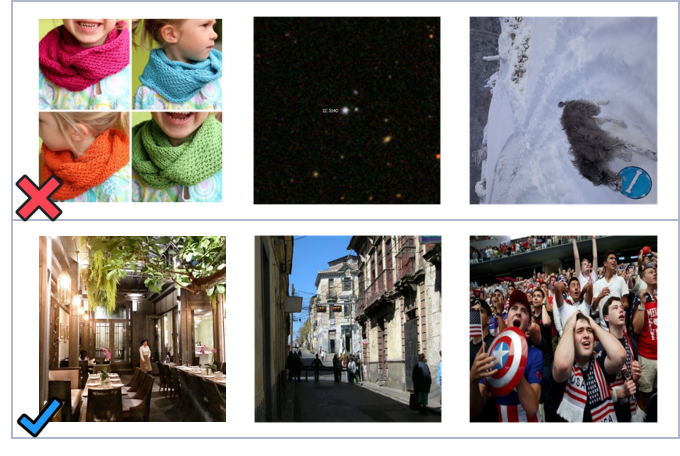}
  \caption{Top: examples of images rejected by expert-defined filtering rules. Bottom: examples of high-quality images that pass the rules.}
  \label{fig:right_false}
\end{figure}

\noindent \textbf{Rules Pool.} Figure~\ref{fig:right_false} shows examples of images that were rejected and accepted based on the filtering rules, the following are the rules defined by experts during the image selection process:
\begin{itemize}
    \item Removing images containing explicit or violent content.
    \item Filtering out blurry, low-resolution, or visually ambiguous images.
    \item Excluding images lacking semantic clarity or spatial structure.
    \item Discarding images with structural ambiguity (e.g., multiple valid puzzle arrangements).
    \item Eliminating misaligned images or those with overly small visual elements after cropping, which hinder spatial reasoning.
\end{itemize}

\noindent \textbf{Task-Specific Template.} The following are detailed templates for each task. Note that <image\_x> denotes a placeholder for the corresponding image input.

\begin{figure}[H]
  \includegraphics[width=\columnwidth]{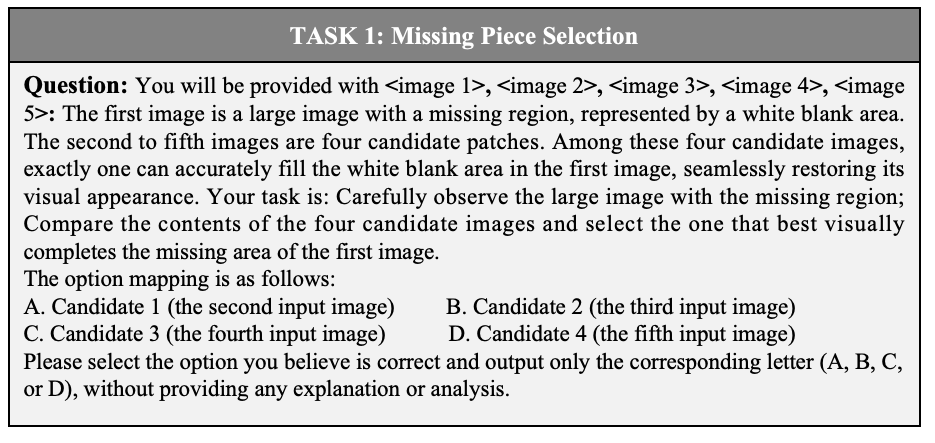}
  \caption{Template of \textit{Missing Piece Selection}, notably, the templates for the Easy and Hard settings are identical.}
  \label{fig:1}
\end{figure}

\begin{figure}[H]
  \includegraphics[width=\columnwidth]{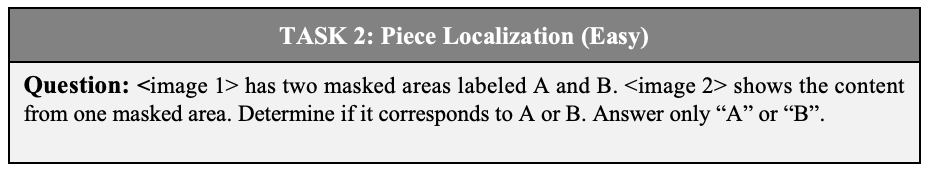}
  \caption{Template of \textit{Piece Localization (Easy)}.}
  \label{fig:2_easy}
\end{figure}

\begin{figure}[H]
  \includegraphics[width=\columnwidth]{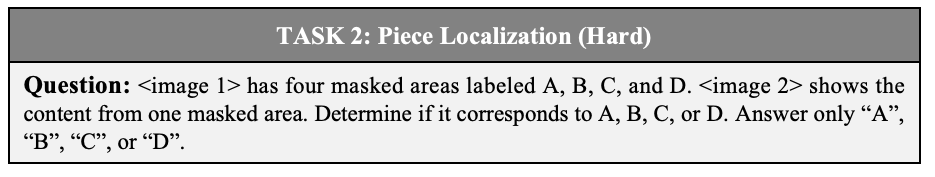}
  \caption{Template of \textit{Piece Localization (Hard)}.}
  \label{fig:2_hard}
\end{figure}

\begin{figure}[H]
  \includegraphics[width=\columnwidth]{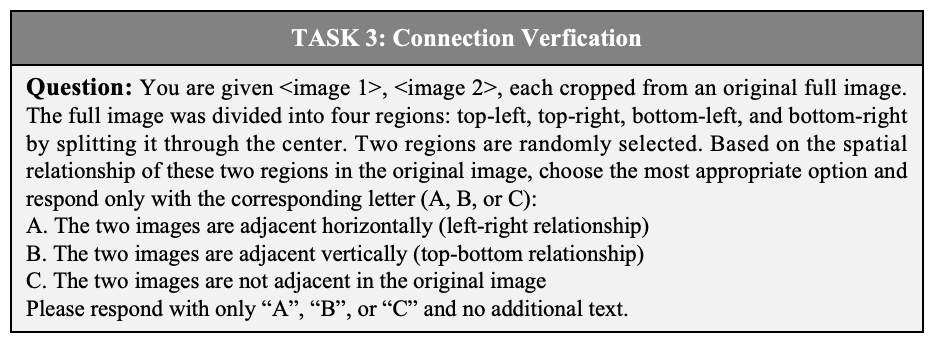}
  \caption{Template of \textit{Connection Verification}.}
  \label{fig:3}
\end{figure}

\begin{figure}[H]
  \includegraphics[width=\columnwidth]{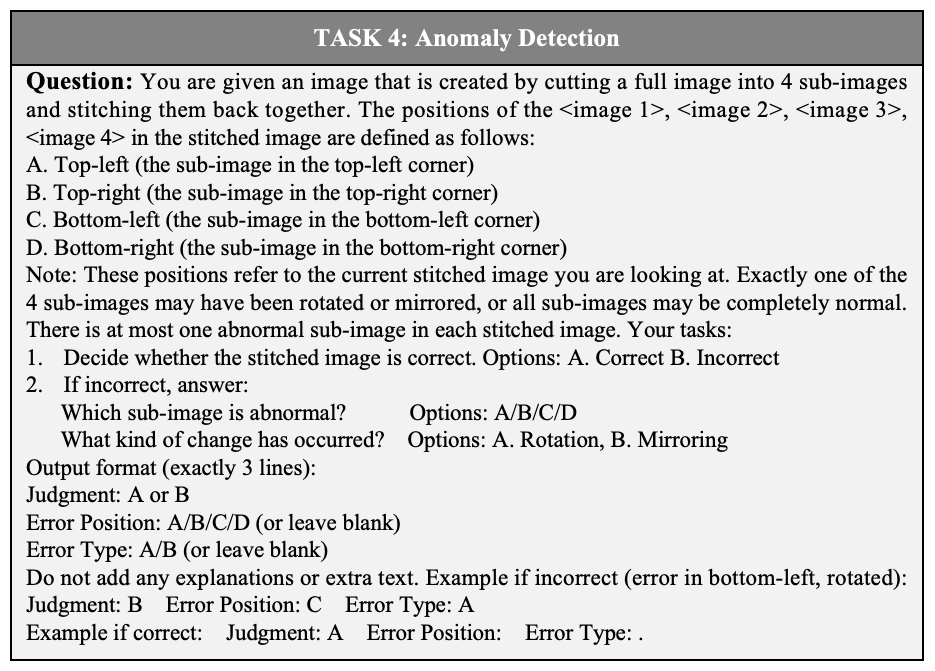}
  \caption{Template of \textit{Anomaly Detection}.}
  \label{fig:4}
\end{figure}

\begin{figure}[H]
  \includegraphics[width=\columnwidth]{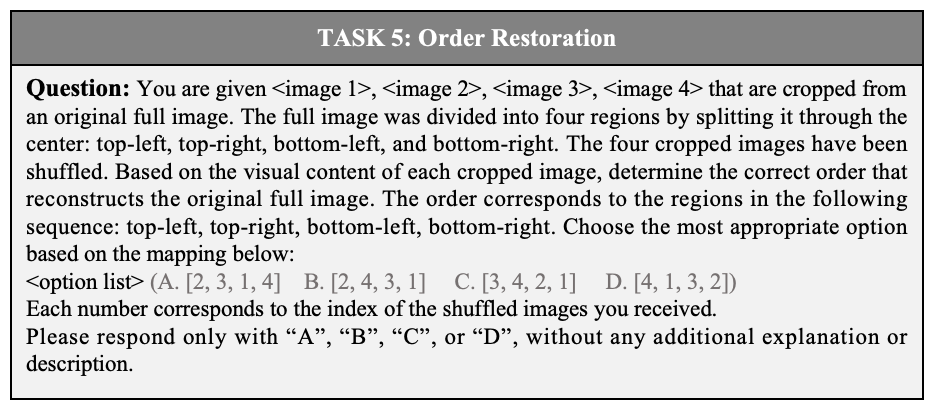}
  \caption{Template of \textit{Order Restoration}. Note: <option list> serves as a placeholder for the answer choices. The text in parentheses is an example and should be removed in actual use. One option is the correct answer, while the remaining three are randomly drawn from the other 23 candidates.}
  \label{fig:5}
\end{figure}

\begin{figure}[H]
  \includegraphics[width=\columnwidth]{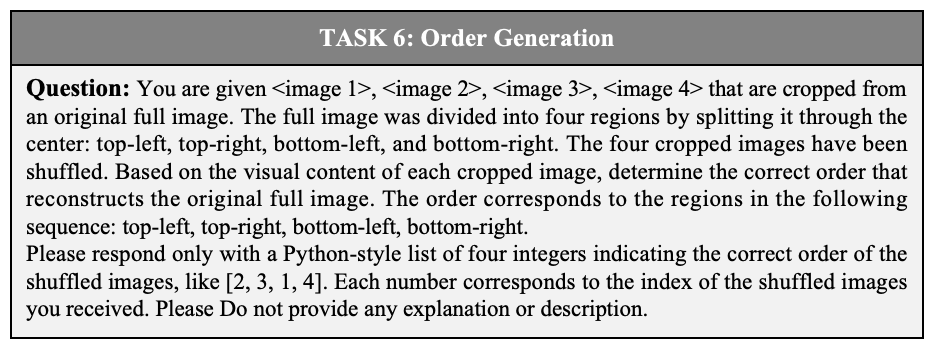}
  \caption{Template of \textit{Order Generation}.}
  \label{fig:6}
\end{figure}

\section{Hardware Setup for Evaluating VLMs.}
\label{sec:hardware}

We evaluate open-source VLMs using hardware configurations scaled to model size. Models with fewer than 20B parameters run on a single NVIDIA A100 80GB GPU. Those between 20B and 40B use two NVIDIA A100 80GB GPUs, while models exceeding 40B are evaluated on four NVIDIA A100 80GB GPUs to meet their greater memory and computational demands.

\end{document}